\pdfoutput=1
\documentclass[11pt,a4paper]{article}
\usepackage[hyperref]{emnlp2020}
\usepackage{times}
\usepackage{graphicx}
\usepackage{caption}
\usepackage{hyperref}
\usepackage{multirow}
\usepackage{CJKutf8}
\usepackage{amssymb}
\usepackage{microtype}

\aclfinalcopy 

\title{Intrinsic Knowledge Evaluation on Chinese Language Models}

\author{Zhiruo Wang \\
	Beijing Normal University \\
	\texttt{zhiruowang@mail.bnu.edu.cn} \\\And
	Renfen Hu \\
	Beijing Normal University \\
	\texttt{irishu@bnu.edu.cn} \\}

\date{}

\begin{document}
\maketitle

\begin{abstract}

Recent NLP tasks have benefited a lot from pre-trained language models (LM) since they are able to encode knowledge of various aspects. However, current LM evaluations focus on downstream performance, hence lack to comprehensively inspect on which aspect and to what extent have they encoded knowledge. This paper addresses both queries by proposing four tasks on syntactic, semantic, commonsense, and factual knowledge, aggregating to a total of $39,308$ questions covering both linguistic and world knowledge in Chinese. Throughout experiments, our probes and knowledge data prove to be a reliable benchmark for evaluating pre-trained Chinese LMs. Our work is publicly available at \href{https://github.com/ZhiruoWang/ChnEval}{https://github.com/ZhiruoWang/ChnEval}

\end{abstract}

\section{Introduction}

Recent years witnessed much success achieved by pre-trained LMs in the field of Natural Language Processing \citep{peters2018deep,devlin2019bert}. The performance of these models is often evaluated on downstream tasks like reading comprehension (RC), natural language inference (NLI), and sentiment analysis (SA). 
However, improvements on downstream hardly explain the reasons behind models' excellence, as well as what they learn during pre-training. Therefore, an emerging body of work starts to investigate the knowledge encoded in their contextual representations.

Linguistic probing methods are designed to uncover the intriguing properties stored in the contextual representations. Among the linguistic knowledge, syntax is broadly explored across sensitive structures \citep{goldberg2019assessing}, grammatical correctness \citep{marvin2018targeted}, and parsing dependencies \citep{hewitt2019structural}.  However, existing language probes face three challenges: (1) A skewing on syntax, for few semantic tasks ever study the contextual representations; (2) Most probes are built as classifiers that require extra training. It raises the question `Do the representations encode linguistic structure, or just that the probe has learned the linguistic task' \citep{hewitt2019designing}? and (3) Existing probing tasks scope to only an English language setting.

\begin{figure}
    \setlength{\belowcaptionskip}{-0.6cm}
    \begin{flushleft}
        \includegraphics[width=.48\textwidth]{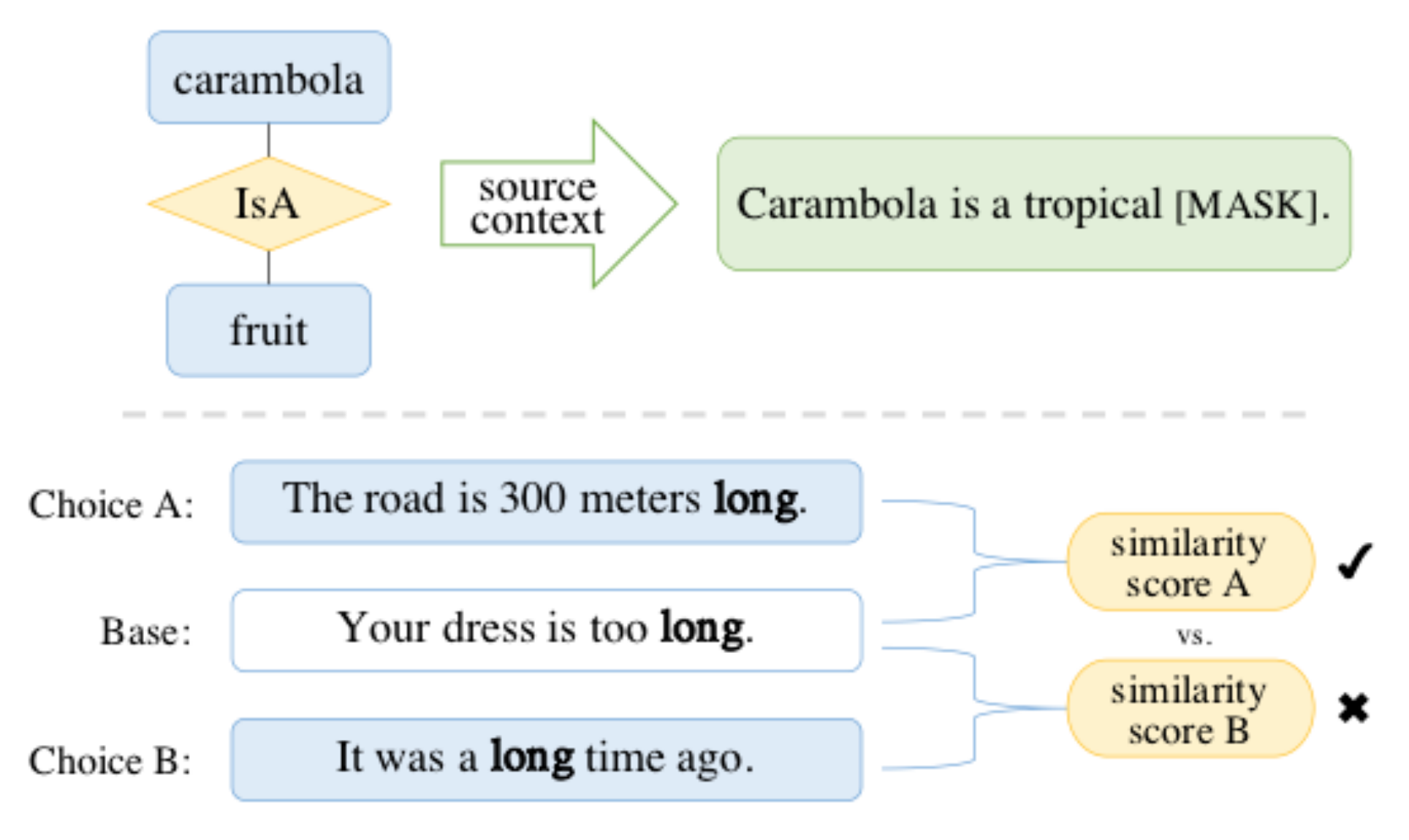}
        \caption{Two forms of tasks.
            Above shows how a knowledge triple is tested in ``fill-in-the-blank'' cloze questions.
            Below is a word-sense similarity task.}
        \label{fig:1}
    \end{flushleft}
\end{figure}

In addition to the linguistics, tasks on common sense and facts are also introduced to test models on memorizing real-world knowledge during pre-training \citep{bisk2019piqa,zhou2019evaluating,petroni2019language}. Nonetheless, the knowledge encoding ability of BERT is controversial \citep{poerner2019bert}, and the template-based cloze questions are often too short to be leveraged by models for informative contextualizations.

Inspired by the above works, this paper proposes the first intrinsic knowledge evaluation benchmark of Chinese pre-trained LMs. 
Linguistically, it covers both the syntactic and semantic knowledge. 
One task aims at the language-specific syntactic features of Chinese, and another on language-independent semantic features.
Meanwhile, we inspect world knowledge from two tasks on common sense and facts, further enable questions with natural contexts. All of the four tasks are designed to fit the LM structures and capabilities, i.e. making predictions directly from deep contextualized embeddings without additional tuning.

In the experiments, we test not only off-the-shelf models from CLUE project \citep{xu2020clue}, but also four BERT variants granted with different training objectives that mimic BERT, RoBERTa, SpanBERT, and ALBERT. Our tasks and data sets prove to constitute a reliable evaluation benchmark. It effectively illustrates the advantages and disadvantages of different LMs over various aspects of knowledge.

\section{Knowledge and Evaluation}

\subsection{Linguistic Knowledge}

Linguistic knowledge is fundamental to language understanding. To examine the linguistic knowledge encoded in pre-trained LMs, we propose two language probing tasks to address both the syntactic and semantic regularities.

\subsubsection{Syntactic Regularities}

Chinese is a typical analytic language without explicit inflections, but uses function words and word order to convey grammatical information \citep{li2018analogical}. In the following case, the auxiliary word \textit{`le'} indicates the perfective tense, and the preposition \textit{`bǎ'} is used to emphasize the object by changing the word order from S-V-O to S-bǎ–O-V:

\textit{wǒ(I) \textbf{bǎ} shū(book) kàn(read) wán(finish) \textbf{le}.}

\textit{(I have finished reading the book.)}

A good word-level test on syntax, hence, is whether they can utilize function words aptly.
This paper considers five categories of function words: conjunctions (C), adverbs (D), prepositions (P), auxiliary words (U), and direction nouns (ND). 
In this task, we mask function words in sentences to form cloze questions, in which the models leverage contextual information to make predictions.



\subsubsection{Semantic Regularities}

As noted by \citet{firth1957synopsis}, `you shall know a word by the company it keeps'. To comprehend a polysemous word, one must dynamically infer its meaning from surrounding contexts.
For example, meanings of the word `\textit{long}' vary in 

A: \textit{The road is \underline{long}.}

B: \textit{I have been exercising for a \underline{long} time.} 

While in A it measures a substantive object, B tells a lapse of time.

Since pre-trained LMs can inherently capture complexities of word use \citep{peters2018deep}, we propose a word sense similarity task to test their discrimination between nuances.
Intuitively, given 

C: \textit{The table is 1-meter \underline{long}.}

A qualified model shall put this `\textit{long}' akin to that of A, and set it apart with B.
As such, this task is built as multiple-choice questions. Each has three sentences---base, answer, and distractor---with identical words. Meaning of target words accords in base-answer and differs in base-distractor. Their final-layer contextual representations $v_{base}$, $v_{answer}$ and $v_{distractor}$ are taken, to compute the cosine similarities of base-answer and base-distractor pairs. We expect base-answer to score higher.

\subsection{World Knowledge}

Besides learning linguistic knowledge, tasks like Question Answering (QA) and Reading Comprehension (RC) often require real-world knowledge beyond contexts. 
We investigate world knowledge in common sense and encyclopedia facts.

\subsubsection{Common Sense}

Common sense is practical judgments about routine affairs. 
For example, `\textit{the hot weather}' makes you`\textit{thirsty}' shows a causality. If a model can sense the common `\textit{thirsty}' from the premise of a hot day, it can benefit from this logical inference.
Following \citet{petroni2019language} that uses ConceptNet \citep{speer2017conceptnet} word pairs, we take the Chinese pairs, put them into the provided text templates \citep{kuo2009community}, and mask items to create clozes.

\subsubsection{Encyclopedia Fact}

Facts, often as (entity, relation, attribute) triples in Knowledge Bases, are helpful in NLP scenarios. 
To answer `\textit{How do carambolas taste like?}', a prior knowledge \textit{(carambola, IsA, fruit)} can inform the possibly sweet or sour taste of fruits.

Different from the template-based commonsense clozes, we present fact triples in their natural contexts (i.e. the source texts that they were extracted from), to allow more contextual information and minimize manual intervention. 
That is, we link the above triple to its wiki introduction\footnote{Only triples fully covered by its wikitext are kept. Paired contents are sheared to lengths within $[16, 128]$ characters.}.
After masking the item `\textit{fruit}', the input reads: 
\textit{Carambola is a tropical [MASK] in Southeast Asia.}
The greater chances that a model predicts words right, the better we assume its encoding of concerned knowledge is.

\begin{table*}
    \centering
    \begin{tabular}{lllccc}
            \hline
            \textbf{Knowledge} & \textbf{Task} & \textbf{Form}  & \textbf{Sub-class} & \textbf{Size}   \\
            \hline
            syntactic & function~word~prediction  & cloze~questions   & 5  & 29345  \\
            semantic & word~sense~similarity  & multiple-choice~questions & N/A  & 5790   \\ 
            \hline
            commonsense        & target~(pair~item)~prediction   & cloze~questions  & N/A   & 3111 \\
            encyclopedia & target~(triple~item)~prediction & cloze~questions & N/A    & 1062   \\
            \hline
    \end{tabular}
    \caption{\label{dataset} An overview of the intrisic evaluation datasets. Refer more details to the Appendix.}
\end{table*}

\subsection{Dataset Construction}

In general, our tasks have two forms: 
(1) multiple-choice questions for semantic knowledge;
and (2) cloze questions for function word, commonsense and fact inferences. Unlike existing probing tasks that use additional classifiers, our tasks allow pre-trained LMs to make predictions directly from their contextual representations. This design mitigates influences of the tuning process \citep{hewitt2019designing}, hence can enable a fairer overall evaluation.

The word sense similarity task uses sentences in CTC corpus\footnote{\href{http://www.aihanyu.org/basic\_v2/index.html}{http://www.aihanyu.org/basic\_v2/index.html}}, a Chinese textbook corpus with manual sense labeling on polysemous words.
To balance the data set, we allow at most $10$ questions per meaning and $50$ questions per word, resulting in $5790$ questions from $372$ words.

For syntactic knowledge, we extract sentences that have function words from CTC corpus as well, in the end created $(3776, 10132, 5887, 5711, 3839)$ clozes for type (C, D, P, U, ND) respectively.


Regarding world knowledge, we present common sense using templates and factual triples in natural contexts just as described above. Target items---prefer object$>$relation (if has)$>$subjects---are masked to create cloze questions. 
Common sense adapts from the Chinese part of ConceptNet \citep{speer2017conceptnet}, in which $3111$ pairs from $13$ relation types are usable after professional manual inspection. 
For facts, we built $1062$ clozes using a Chinese Knowledge Base built upon encyclopedia called CN-DBpedia \citep{xu2017cn}.

In these data sets, target words only appear once in their contexts, and sentences within a data set never repeat. We end up obtained four data sets counting to $39,308$ questions in total. Table \ref{dataset} is a summary of these knowledge data sets. See more details of manual checks and illustrated examples in Appendix B and C.
For evaluation results,  unless otherwise stated, we report the average score if multiple classes occur\footnote{Calculate prediction accuracy firstly within each category, then averaged across classes.}.

\begin{table*}
    \centering
    \resizebox{\textwidth}{!}{
    \begin{tabular}{llcccc}
        \hline
        \textbf{Data Set} & \textbf{Metrics} & \textbf{BERT} & \textbf{BERT} & \textbf{BERT} & \textbf{RoBERTa} \\
        \small{\textbf{}} & \small{\textbf{}} & \small{\textbf{}} & \small{\textbf{-wwm}} & \small{\textbf{-wwm-ext}} & \small{\textbf{-wwm-ext}} \\
        \hline
        {syntactic} & {P@1/10} & {38.8 / 76.8} & {42.7 / 77.8} & {42.4 / 77.5} & \textbf{56.9 / 88.0} \\
        {semantic} & {acc.} & {69.7} & {69.8} & {71.2} & \textbf{73.1} \\
        \hline
        {commonsense} & {P@1/10} & {3.38 / 21.63} & {1.32 / 18.55} & {2.12 / 15.30} & \textbf{19.83 / 43.56} \\
        {encyclopedia} & {P@1/10} & {29.1 / 65.1} & {34.8 / 67.7} & {32.6 / 68.9} & \textbf{60.3 / 85.7} \\
        
        \hline
        {CMRC \citep{cui-emnlp2019-cmrc2018}} & {avg. EM/F1} & {68.7 / 86.3} & {69.1 / 86.7} & {70.0 / 87.0} & \textbf{71.4 / 88.8} \\
        {XNLI \citep{conneau2018xnli}} & {avg. acc.} & {77.5} & {78.0} & {78.3} & \textbf{78.3} \\
        {ChnSentiCorp \citep{chinese-bert-wwm}} & {avg. acc.} & {94.7} & {\textbf{95.0}} & {94.7} & {94.8} \\
        {THUCNews \citep{sum2016thuctc}} & {avg. acc.} & \textbf{97.6} & \textbf{97.6} & {97.5} & {97.5} \\
        \hline
    \end{tabular} }
    \caption{\label{sota-result} Off-the-shelf results on intrinsic and extrinsic tasks. For cloze tests, we report top-1 and top-10 precision as P@1 and P@10.  CMRC, XNLI, ChnSentiCorp, THUCNews test Reading Comprehension, cross-lingual Natural Language Inference, Sentiment Analysis and Document Classification respectively.    }
\end{table*}

\begin{table*}
    \centering
    \setlength{\belowcaptionskip}{-0.5cm}
    \begin{tabular}{llcccc}
        \hline
        \textbf{Data Set} & \textbf{Metrics} & \textbf{MLM} & \textbf{MLM + SBO} & \textbf{MLM + SOP} & \textbf{MLM + NSP} \\
        \hline
        {syntactic} & {P@1/10} & \textbf{53.4 / 86.4} & {41.4 / 78.1} & {36.3 / 72.7} & {50.0 / 83.8} \\
        {semantic} & {acc.} & \textbf{73.4} & {69.0} & {70.1} & {71.6} \\
        \hline
        {commonsense} & {P@1/10} & \textbf{13.95 / 37.93} & {7.88 / 22.50} & {3.44 / 16.33} & {9.39 / 37.39} \\
        {encyclopedia} & {P@1/10} & \textbf{69.5 / 90.7} & {48.3 / 80.2} & {33.7 / 69.1} & {63.6 / 85.0} \\
        
        
        \hline
    \end{tabular}
    \caption{\label{objective-result} Knowledge evaluation results on objective-variants. }
\end{table*}

\section{Experiment}

\subsection{Models}

Many Chinese language models are publicly available at CLUE \citep{xu2020clue}. To reduce computing cost and keep candidates comparable in size, we test the models in BASE-size: BERT \citep{devlin2019bert}, BERT-wwm, BERT-wwm-ext, and RoBERTa-wwm-ext \citep{chinese-bert-wwm}\footnote{More specified model configurations list in Appendix C}.

Also, to compare the performance between training objectives, we implement four variants:

\textbf{BERT (MLM+NSP)}: Masked Language Model and Next Sentence Prediction as in BERT.

\textbf{RoBERTa (MLM)}: Remove NSP, often benefit from longer sequences and fewer topic conflicts.

\textbf{SpanBERT (MLM+SBO)}: Add a span-shrunk version of the Subject Boundary Objective \citep{joshi2020spanbert}. We mask single tokens instead of spans to control variables.

\textbf{ALBERT (MLM+SOP)}: Replace NSP with the Sentence Order Prediction \citep{lan2019albert}, guide bi-spans on inter-logic rather than topic conformity.

Training from scratch costs, so we initialize them with BERT-base-chinese and train for $500,000$ additional steps at a $32$ batch size using Baidu Baike corpus. Other settings align with BERT.

\subsection{Results and discussion}


Table \ref{sota-result} shows the results of off-the-shelf models, on our intrinsic knowledge tasks and four extrinsic NLP tasks. Our three observations read as follows.

\begin{figure}
    \begin{flushleft}
    \setlength{\belowcaptionskip}{-0.8cm}
        \includegraphics[width=.45\textwidth, height=0.18\textheight]{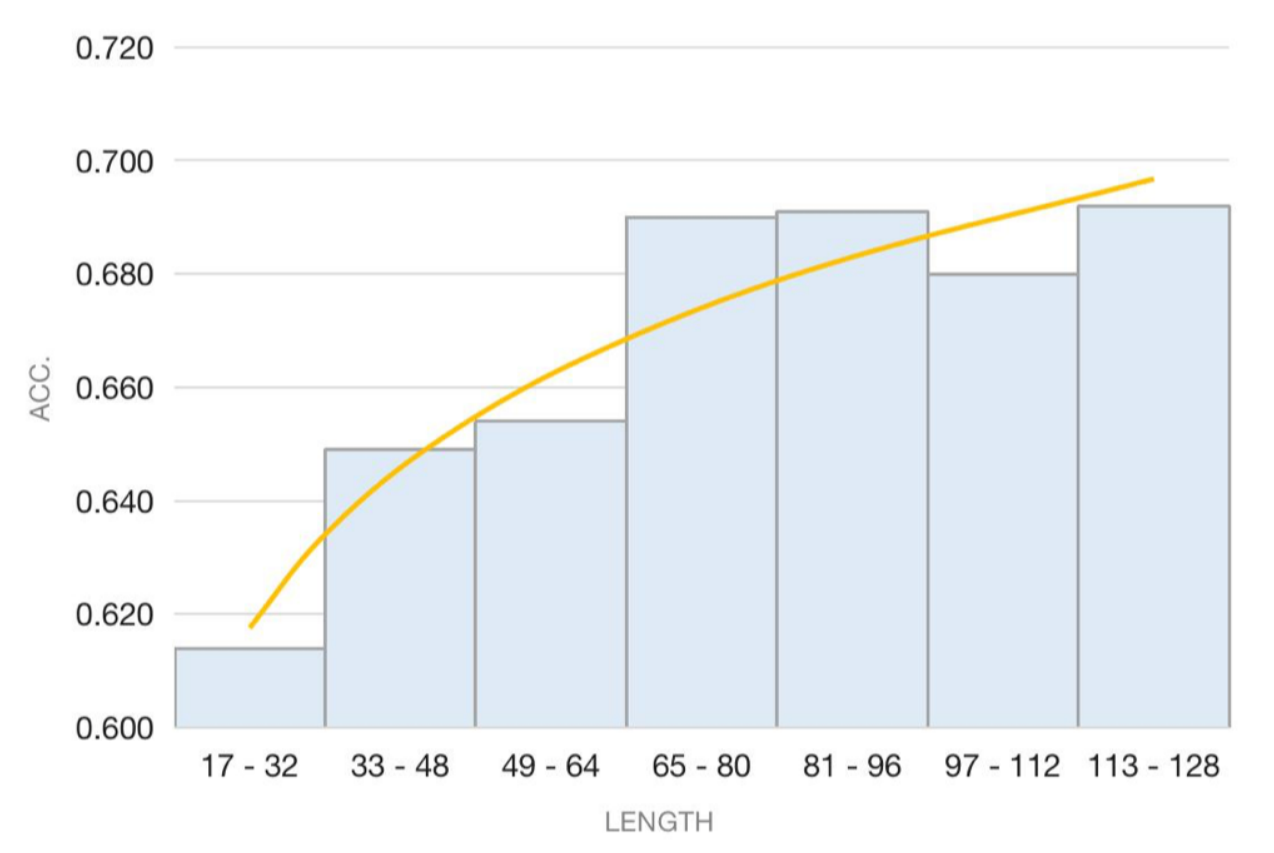}
        \caption{Bucket-averaged accuracies on fact clozes. Accuracy grows by increasing length.}
        \label{fig:2}
    \end{flushleft}
\end{figure}

\textbf{First}, pre-trained Chinese LMs using natural contexts capture the linguistic and factual knowledge well. However, they do poorly on commonsense questions, probably because they are not fully capable of storing relationally-structured knowledge \citep{poerner2019bert}. For another, template-based clozes might be too short triggers for models to yield informative contextual representations. To further verify our hypothesis, we bucket fact clozes based on text length, study the trend depending on lengths, then illustrate in Figure \ref{fig:2}.

\textbf{Second}, pre-process with whole-word-masking and pre-train with additional data, consistently enhance the performance in most intrinsic tasks. Removing the NSP objective further helps. RoBERTa-wwm-ext, who integrates all of the three advantages, scores the highest on intrinsic tasks.

\textbf{Third}, by comparing the intrinsic tasks against extrinsic ones, we observe that the intrinsic tasks ring more sensitive to changes in model structures and training data. Among extrinsic tasks, only in Reading Comprehension that models consistently improve with upgraded masking strategy and training corpora. XNLI doesn't lead to any difference between BERT-wwm-ext and RoBERTa-wwm-ext. The other two classification tasks vary trivially across four models. These results suggest that our intrinsic tasks can reflect model discrepancies in more elaborate ways than extrinsic ones, further, unveil knowledge encoding from various aspects.


Note that these intrinsic knowledge evaluations can also shed light on the structural design of LMs.
Since most off-the-shelf models vary diversely (in the corpus, parameters, and more), we focus on a single factor, the training objectives. 
We implemented four BERT variants and make a preliminary study of their effects on knowledge encoding abilities.
As shown in Table \ref{objective-result}. MLM (RoBERTa) strikes the best in all knowledge aspects. Boundary information (SBO) barely helps, for it may suit spans better than single tokens. NSP surpasses SOP in most cases, showing a priority of topic conformity at bi-span training.

\section{Conclusion and Future Work}

In this paper, we present the first intrinsic knowledge evaluation data set of Chinese pre-trained LMs, ranging from syntactic, semantic, commonsense, to factual knowledge.
The experiments show that our tasks and data sets constitute a reliable evaluation benchmark. It effectively uncovers not only the pros and cons of different LMs over a varied aspects of knowledge. Further, it offers insight on structural designs. With these tasks, we can make an in-depth analysis of LM knowledge encoding in the future and better understand the ``black box'' of Neural Network methods.
Last but not least, our task building methods can apply well to other language environments.

\bibliographystyle{acl_natbib}
\bibliography{emnlp2020}

\begin{thebibliography}{22}
\expandafter\ifx\csname natexlab\endcsname\relax\def\natexlab#1{#1}\fi

\bibitem[{Bisk et~al.(2019)Bisk, Zellers, Bras, Gao, and Choi}]{bisk2019piqa}
Yonatan Bisk, Rowan Zellers, Ronan~Le Bras, Jianfeng Gao, and Yejin Choi. 2019.
\newblock Piqa: Reasoning about physical commonsense in natural language.
\newblock \emph{arXiv preprint arXiv:1911.11641}.

\bibitem[{Conneau et~al.(2018)Conneau, Rinott, Lample, Williams, Bowman,
  Schwenk, and Stoyanov}]{conneau2018xnli}
Alexis Conneau, Ruty Rinott, Guillaume Lample, Adina Williams, Samuel~R.
  Bowman, Holger Schwenk, and Veselin Stoyanov. 2018.
\newblock Xnli: Evaluating cross-lingual sentence representations.
\newblock In \emph{Proceedings of the 2018 Conference on Empirical Methods in
  Natural Language Processing}. Association for Computational Linguistics.

\bibitem[{Cui et~al.(2019{\natexlab{a}})Cui, Che, Liu, Qin, Yang, Wang, and
  Hu}]{chinese-bert-wwm}
Yiming Cui, Wanxiang Che, Ting Liu, Bing Qin, Ziqing Yang, Shijin Wang, and
  Guoping Hu. 2019{\natexlab{a}}.
\newblock Pre-training with whole word masking for chinese bert.
\newblock \emph{arXiv preprint arXiv:1906.08101}.

\bibitem[{Cui et~al.(2019{\natexlab{b}})Cui, Liu, Che, Xiao, Chen, Ma, Wang,
  and Hu}]{cui-emnlp2019-cmrc2018}
Yiming Cui, Ting Liu, Wanxiang Che, Li~Xiao, Zhipeng Chen, Wentao Ma, Shijin
  Wang, and Guoping Hu. 2019{\natexlab{b}}.
\newblock \href {https://doi.org/10.18653/v1/D19-1600} {A span-extraction
  dataset for {C}hinese machine reading comprehension}.
\newblock In \emph{Proceedings of the 2019 Conference on Empirical Methods in
  Natural Language Processing and the 9th International Joint Conference on
  Natural Language Processing (EMNLP-IJCNLP)}, pages 5886--5891, Hong Kong,
  China. Association for Computational Linguistics.

\bibitem[{Devlin et~al.(2019)Devlin, Chang, Lee, and
  Toutanova}]{devlin2019bert}
Jacob Devlin, Ming-Wei Chang, Kenton Lee, and Kristina Toutanova. 2019.
\newblock Bert: Pre-training of deep bidirectional transformers for language
  understanding.
\newblock In \emph{Proceedings of the 2019 Conference of the North American
  Chapter of the Association for Computational Linguistics: Human Language
  Technologies, Volume 1 (Long and Short Papers)}, pages 4171--4186.

\bibitem[{Firth(1957)}]{firth1957synopsis}
John~R Firth. 1957.
\newblock A synopsis of linguistic theory, 1930-1955.
\newblock \emph{Studies in linguistic analysis}.

\bibitem[{Goldberg(2019)}]{goldberg2019assessing}
Yoav Goldberg. 2019.
\newblock Assessing bert's syntactic abilities.
\newblock \emph{arXiv preprint arXiv:1901.05287}.

\bibitem[{Hewitt and Liang(2019)}]{hewitt2019designing}
John Hewitt and Percy Liang. 2019.
\newblock Designing and interpreting probes with control tasks.
\newblock \emph{arXiv preprint arXiv:1909.03368}.

\bibitem[{Hewitt and Manning(2019)}]{hewitt2019structural}
John Hewitt and Christopher~D Manning. 2019.
\newblock A structural probe for finding syntax in word representations.
\newblock In \emph{Proceedings of the 2019 Conference of the North American
  Chapter of the Association for Computational Linguistics: Human Language
  Technologies, Volume 1 (Long and Short Papers)}, pages 4129--4138.

\bibitem[{Joshi et~al.(2020)Joshi, Chen, Liu, Weld, Zettlemoyer, and
  Levy}]{joshi2020spanbert}
Mandar Joshi, Danqi Chen, Yinhan Liu, Daniel~S Weld, Luke Zettlemoyer, and Omer
  Levy. 2020.
\newblock Spanbert: Improving pre-training by representing and predicting
  spans.
\newblock \emph{Transactions of the Association for Computational Linguistics},
  8:64--77.

\bibitem[{Kuo et~al.(2009)Kuo, Lee, Chiang, Wang, Shen, Chan, and
  Hsu}]{kuo2009community}
Yen-ling Kuo, Jong-Chuan Lee, Kai-yang Chiang, Rex Wang, Edward Shen, Cheng-wei
  Chan, and Jane Yung-jen Hsu. 2009.
\newblock Community-based game design: experiments on social games for
  commonsense data collection.
\newblock In \emph{Proceedings of the acm sigkdd workshop on human
  computation}, pages 15--22.

\bibitem[{Lan et~al.(2019)Lan, Chen, Goodman, Gimpel, Sharma, and
  Soricut}]{lan2019albert}
Zhenzhong Lan, Mingda Chen, Sebastian Goodman, Kevin Gimpel, Piyush Sharma, and
  Radu Soricut. 2019.
\newblock Albert: A lite bert for self-supervised learning of language
  representations.
\newblock \emph{arXiv preprint arXiv:1909.11942}.

\bibitem[{Li et~al.(2018)Li, Zhao, Hu, Li, Liu, and Du}]{li2018analogical}
Shen Li, Zhe Zhao, Renfen Hu, Wensi Li, Tao Liu, and Xiaoyong Du. 2018.
\newblock Analogical reasoning on chinese morphological and semantic relations.
\newblock In \emph{Proceedings of the 56th Annual Meeting of the Association
  for Computational Linguistics (Volume 2: Short Papers)}, pages 138--143.

\bibitem[{Marvin and Linzen(2018)}]{marvin2018targeted}
Rebecca Marvin and Tal Linzen. 2018.
\newblock Targeted syntactic evaluation of language models.
\newblock \emph{arXiv preprint arXiv:1808.09031}.

\bibitem[{Peters et~al.(2018)Peters, Neumann, Iyyer, Gardner, Clark, Lee, and
  Zettlemoyer}]{peters2018deep}
Matthew~E Peters, Mark Neumann, Mohit Iyyer, Matt Gardner, Christopher Clark,
  Kenton Lee, and Luke Zettlemoyer. 2018.
\newblock Deep contextualized word representations.
\newblock \emph{arXiv preprint arXiv:1802.05365}.

\bibitem[{Petroni et~al.(2019)Petroni, Rockt{\"a}schel, Lewis, Bakhtin, Wu,
  Miller, and Riedel}]{petroni2019language}
Fabio Petroni, Tim Rockt{\"a}schel, Patrick Lewis, Anton Bakhtin, Yuxiang Wu,
  Alexander~H Miller, and Sebastian Riedel. 2019.
\newblock Language models as knowledge bases?
\newblock \emph{arXiv preprint arXiv:1909.01066}.

\bibitem[{Poerner et~al.(2019)Poerner, Waltinger, and
  Sch{\"u}tze}]{poerner2019bert}
Nina Poerner, Ulli Waltinger, and Hinrich Sch{\"u}tze. 2019.
\newblock Bert is not a knowledge base (yet): Factual knowledge vs. name-based
  reasoning in unsupervised qa.
\newblock \emph{arXiv preprint arXiv:1911.03681}.

\bibitem[{Speer et~al.(2017)Speer, Chin, and Havasi}]{speer2017conceptnet}
Robyn Speer, Joshua Chin, and Catherine Havasi. 2017.
\newblock Conceptnet 5.5: An open multilingual graph of general knowledge.
\newblock In \emph{Thirty-First AAAI Conference on Artificial Intelligence}.

\bibitem[{Sum et~al.(2016)Sum, Li, Guo, Zhao, Zheng, Si, and
  Liu}]{sum2016thuctc}
M~Sum, J~Li, Z~Guo, Y~Zhao, Y~Zheng, X~Si, and Z~Liu. 2016.
\newblock Thuctc: an efficient chinese text classifier.
\newblock \emph{GitHub Repository}.

\bibitem[{Xu et~al.(2017)Xu, Xu, Liang, Xie, Liang, Cui, and Xiao}]{xu2017cn}
Bo~Xu, Yong Xu, Jiaqing Liang, Chenhao Xie, Bin Liang, Wanyun Cui, and Yanghua
  Xiao. 2017.
\newblock Cn-dbpedia: A never-ending chinese knowledge extraction system.
\newblock In \emph{International Conference on Industrial, Engineering and
  Other Applications of Applied Intelligent Systems}, pages 428--438. Springer.

\bibitem[{Xu et~al.(2020)Xu, Zhang, Li, Hu, Cao, Liu, Li, Li, Sun, Xu
  et~al.}]{xu2020clue}
Liang Xu, Xuanwei Zhang, Lu~Li, Hai Hu, Chenjie Cao, Weitang Liu, Junyi Li,
  Yudong Li, Kai Sun, Yechen Xu, et~al. 2020.
\newblock Clue: A chinese language understanding evaluation benchmark.
\newblock \emph{arXiv preprint arXiv:2004.05986}.

\bibitem[{Zhou et~al.(2019)Zhou, Zhang, Cui, and Huang}]{zhou2019evaluating}
Xuhui Zhou, Yue Zhang, Leyang Cui, and Dandan Huang. 2019.
\newblock Evaluating commonsense in pre-trained language models.
\newblock \emph{arXiv preprint arXiv:1911.11931}.

\end{thebibliography}

\newpage
\appendix

\section{Model Specifications} \label{sec:model}
Here we specify several configuration details about off-the-shelf candidates in Table \ref{sota-config}.

\begin{CJK*}{UTF8}{gbsn}
	\begin{table*}
		\centering
		\resizebox{\textwidth}{!}{
			\begin{tabular}{|l|l|l|l|l|}
				\hline
				- & \textbf{Masking} & \textbf{Data~Source} & \textbf{Training~Steps} & \textbf{Optimizer} \\
				\hline
				BERT & WordPiece & wiki & $1M^{MAX512}$ & AdamW \\
				\hline
				BERT-wwm & WWM & wiki & $100K^{MAX128}~+~100K^{MAX512}$ & LAMB \\
				\hline
				BERT-wwm-ext & WWM & wiki+ext & $1M^{MAX128}~+~400K^{MAX512}$ & LAMB \\
				\hline
				RoBERTa-wwm-ext & WWM & wiki+ext & $1M^{MAX512}$ & AdamW \\
				\hline     
			\end{tabular}
		}
		\caption{\label{sota-config} Varied configurations of off-the-shelf pre-trained language models. }
	\end{table*}
\end{CJK*}


\section{Data Sets}
\noindent
Chinese Knowledge Evaluation (CKE) benchmark includes four tasks on linguistic and world knowledge. For a better illustration, a summary of the data sets is shown in Table \ref{a-summary}.
\vspace{0.3cm}

\begin{table*}
	\resizebox{\textwidth}{!}{
		\begin{tabular}{|c|c|c|c|c|r|} 
			\hline
			\textbf{Knowledge} & \textbf{Data~Set} & \textbf{Task} & \textbf{Form} & \textbf{Sub-class} & \textbf{Size} \\ 
			\hline
			\multirow{6}{*}{linguistic} & \multirow{5}{*}{syntactic} & \multirow{5}{*}{word~function~prediction} & \multirow{5}{*}{cloze~questions} & conjunction~(C) & 3776 \\ 
			\cline{5-6}
			&   &   &  & adverb~(D) & 10132 \\ 
			\cline{5-6}
			&   &   &  & preposition~(P) & 5887 \\ 
			\cline{5-6}
			&   &   &  & auxiliary~(U) & 5711 \\ 
			\cline{5-6}
			&   &   &  & direction~nouns~(ND) & 3839 \\ 
			\cline{2-6}
			& semantic & word-sense~similarity & multiple-choice~questions & N/A & 5790 \\ 
			\hline
			\multirow{2}{*}{world} & commonsense & {target~(pair~item)~prediction} & {cloze~questions} & N/A & 3111 \\ 
			\cline{2-6}
			& encyclopedia & target~(triple~item)~prediction & cloze~questions & N/A & 1062 \\
			\hline
		\end{tabular}
	}
	\caption{\label{a-summary} A summary on Chinese Knowledge Evaluation data sets. }
\end{table*}

\section{Manual Check} \label{sec:check}
The source of linguistic knowledge has been repetitively checked by professionals. Encyclopedia facts are supported with references and open to user reviews. Hence, their resulting knowledge data sets have ensured qualities.
Similarly, to ensure the quality of commonsense data, a series of manual revisions are performed on ConceptNet (the Chinese part) by six graduate students of linguistics majors. Also, to ensure a unified common sense and language sense, training and annotation trials are performed before the revision.
Revisions on the Chinese ConceptNet pairs include:

\textbf{Step 1}: Cases having non-unique answers are removed.

\textbf{Step 2}: Cases not in line with human commonsense are manually filtered out. They can be categorized into three types:

\noindent $\bullet~$\textbf{illogical or perverse}: (relation: MotivatedByGoal)

\begin{CJK*}{UTF8}{gbsn} 你会[?]因为你没钱。[哭]\end{CJK*}  \\(You will [?] because you have no money. [cry]) 

\noindent $\bullet~$\textbf{indefinite answer}: (relation: HasSubevent)

\begin{CJK*}{UTF8}{gbsn} 可能代表一种元素。[钠]\end{CJK*}  \\([?] may represent an element. [sodium])

\noindent $\bullet~$\textbf{violate universal value}: (relation: Desires)

\begin{CJK*}{UTF8}{gbsn} [?]惧怕庙宇。[鬼]\end{CJK*}  \\([?] fear temple. [ghost])

\textbf{Step 3}: For cases that conform to common sense but are ungrammatical, sentences are further manually modified or re-written. For an example of the relation type 'Causes'(though the English translation may hardly make sense):

\begin{CJK*}{UTF8}{gbsn} 镜子会让你[照]。\end{CJK*} \\(The mirror will let you take a look.) 

changes to: \begin{CJK*}{UTF8}{gbsn} 镜子是用来[照]的。 \end{CJK*} \\(The mirror is used for taking a look.)

Note that after modifying the ungrammatical parts, the original relation type of that sentence may not be retained, such as the `Causes' changes to `UsedFor' in the above example. Therefore, the final dataset no longer divides into different relation types but combines into a single file.

\textbf{Step 4}: Finally, manually proofread the results of \textbf{Step 2} and \textbf{Step 3}.

\section{Examples} \label{sec:example}
\noindent	
In this section, we illustrate several examples for each of the four introduced tasks.
\paragraph{Syntactic Regularities}
For the syntax of words, we showcase one function word for each function class in Table \ref{word-syntax}.

\begin{CJK*}{UTF8}{gbsn}
\begin{table*}
	\centering
	\resizebox{\textwidth}{!}{
		\begin{tabular}{|c|l|l|} 
			\hline
			\textbf{Function} & \multicolumn{1}{c|}{\textbf{Word}} & \multicolumn{1}{c|}{\textbf{Sentence}} \\ 
			\hline
			\multirow{2}{*}{conjunction~(C)} & 但 & 我会跳舞，\textbf{但}跳得不怎么样。\\ 
			& $d\grave{a}n$ & I~can~dance,~\textbf{but}~not~very~well. \\ 
			\hline
			\multirow{2}{*}{adverb~(D)} & 很 & 我在青岛住过三年，\textbf{很}喜爱它。\\ 
			& $h\breve{e}n$ & I~have~lived~in~Qingdao~for~three~years~and~love~it~\textbf{very~much}. \\ 
			\hline
			\multirow{2}{*}{preposition~(P)} & 被 & 你看，花瓶也\textbf{被}他们打破了。\\ 
			& $b\grave{e}i$ & Look,~the~vase~was~also~broken~by~them. \\ 
			\hline
			\multirow{2}{*}{auxiliary~(U)} & 吗 & 知道是什么意思\textbf{吗}？\\ 
			& $m\bar{a}$ & \textbf{Do~you}~know~what~it~means? \\ 
			\hline
			\multirow{2}{*}{direction~nouns~(ND)} & 里 & 小狗怎么在厨房\textbf{里}叫呢？\\ 
			& $l\breve{i}$ & Why~do~puppies~bark~\textbf{in}~the~kitchen? \\
			\hline
		\end{tabular}
	}
	\caption{\label{word-syntax} Example cloze questions from each functional word category. We replace bold words with [MASK] before input.}
\end{table*}
\end{CJK*}

\paragraph{Semantic Regularities}
For the meaning of words, we take the Chinese word `\begin{CJK*}{UTF8}{gbsn}兜\end{CJK*}' ($d\bar{o}u$) as an example, Table \ref{word-mean-candidate} the candidate sentences that are extracted from the textbook corpus, and Table \ref{word-mean-question} lists two example questions that are generated from the candidates.
	
\begin{CJK*}{UTF8}{gbsn}
	\begin{table*}
		\centering
		\begin{tabular}{|c|c|l|} 
			\hline
			\textbf{Word} & \textbf{Sense ID} & \multicolumn{1}{c|}{\textbf{Candidate~Sentences}} \\ 
			\hline
			\multirow{12}{*}{兜} & \multirow{6}{*}{0} & 连小学生也有手机，只要装在衣服\textbf{兜}儿里就可以。 \\ 
			&  & \begin{tabular}[c]{@{}l@{}}Even~elementary~school~students~have~cell~phones,~\\as~long~as~they~are~in~a~\textbf{pocket}~of~clothes.\end{tabular} \\ 
			\cline{3-3}
			&  & 推让了半天，最后我还是把钱塞进了他的\textbf{兜}里。 \\ 
			&  & After~a~long~time,~I~still~put~the~money~in~his~\textbf{pocket}. \\ 
			\cline{3-3}
			&  & 母亲实在太想孙子了，进屋就从\textbf{兜}儿里掏出一把糖来给孙子。 \\ 
			&  & \begin{tabular}[c]{@{}l@{}}My~mother~really~missed~her~grandson.~\\She~took~a~handful~of~sugar~from~her~\textbf{pocket}~to~give~her~grandson.\end{tabular} \\ 
			\cline{2-3}
			& \multirow{2}{*}{1} & 车夫回来的时候\textbf{兜}不到生意。 \\ 
			&  & The~driver~couldn't~\textbf{take}~business~when~he~came~back. \\ 
			\cline{2-3}
			& \multirow{4}{*}{2} & 我喜欢从一条熟的道路出去溜达，然后从一条生的道路\textbf{兜}个圈子回家。\\ 
			&  & \begin{tabular}[c]{@{}l@{}}I~like~to~walk~out~from~a~familiar~road,~\\and~then~go~home~\textbf{in}~a~circle~from~a~raw~road.\end{tabular} \\ 
			\cline{3-3}
			&  & 假如我显露出困惑，老师就会停顿他讲解的步伐，在原地连\textbf{兜}几个圈子。\\ 
			&  & \begin{tabular}[c]{@{}l@{}}If~I~show~confusion,~the~teacher~will~stop~the~pace~of~his~explanations,~\\and~\textbf{make}~a~few~circles~in~the~same~place.\end{tabular}  \\
			\hline
		\end{tabular}
		\caption{\label{word-mean-candidate} An example poly-semantic Chinese word conveying multiple meanings in different sentences. }
	\end{table*}
\end{CJK*}

\begin{table*}
	\centering
	\begin{tabular}{|c|c|l|} 
		\hline
		\textbf{No.} & \textbf{Type} & \multicolumn{1}{c|}{\textbf{Sentences}} \\ 
		\hline
		\multirow{6}{*}{0} & \multirow{2}{*}{Base} & \begin{CJK*}{UTF8}{gbsn} 我喜欢从一条熟的道路出去溜达，然后从一条生的道路\textbf{兜}个圈子回家。\end{CJK*}\\ 
		&  & \begin{tabular}[c]{@{}l@{}}I~like~to~walk~out~from~a~familiar~road,~\\and~then~go~home~\textbf{in}~a~circle~from~a~raw~road.\end{tabular} \\ 
		\cline{2-3}
		& \multirow{2}{*}{Answer} & \begin{CJK*}{UTF8}{gbsn} 假如我显露出困惑，老师就会停顿他讲解的步伐，在原地连\textbf{兜}几个圈子。\end{CJK*}\\ 
		&  & \begin{tabular}[c]{@{}l@{}}If~I~show~confusion,~the~teacher~will~stop~the~pace~of~his~explanations,~\\and~\textbf{make}~a~few~circles~in~the~same~place.\end{tabular}  \\ 
		\cline{2-3}
		& \multirow{2}{*}{Distractor} & \begin{CJK*}{UTF8}{gbsn} 连小学生也有手机，只要装在衣服\textbf{兜}儿里就可以。\end{CJK*}\\ 
		&  & \begin{tabular}[c]{@{}l@{}}Even~elementary~school~students~have~cell~phones,~\\as~long~as~they~are~in~a~\textbf{pocket}~of~clothes.\end{tabular} \\ 
		\hline
		\multirow{6}{*}{1} & \multirow{2}{*}{Base} & \begin{CJK*}{UTF8}{gbsn} 推让了半天，最后我还是把钱塞进了他的\textbf{兜}里。\end{CJK*} \\ 
		&  & After~a~long~time,~I~still~put~the~money~in~his~\textbf{pocket}. \\ 
		\cline{2-3}
		& \multirow{2}{*}{Answer} & \begin{CJK*}{UTF8}{gbsn} 母亲实在太想孙子了，进屋就从\textbf{兜}儿里掏出一把糖来给孙子。\end{CJK*}\\ 
		&  & \begin{tabular}[c]{@{}l@{}}My~mother~really~missed~her~grandson.~\\She~took~a~handful~of~sugar~from~her~\textbf{pocket}~to~give~her~grandson.\end{tabular} \\ 
		\cline{2-3}
		& \multirow{2}{*}{Distractor} & \begin{CJK*}{UTF8}{gbsn} 车夫回来的时候\textbf{兜}不到生意。\end{CJK*}\\ 
		&  & The~driver~couldn't~\textbf{take}~business~when~he~came~back. \\
		\hline
	\end{tabular}
	\caption{\label{word-mean-question} Example multiple-choice questions create from collected sentences. }
\end{table*}
\paragraph{Common Sense}
We examplify word pairs and text templates for each relation in Table \ref{common-sense}. 

\begin{CJK*}{UTF8}{gbsn}
	\begin{table*}
		\centering
		\begin{tabular}{|c|c|l|} 
			\hline
			\textbf{Subject} & \textbf{Object} & \textbf{Text~Template} \\
			\hline
			床 & 卧室 & \textbf{床}在卧室里。\\
			bed & bedroom & \textbf{Bed}~is~in~the~bedroom. \\ 
			\hline
			悲伤 & 哭 & 悲伤的时候，你会\textbf{哭}。\\
			sad & cry & When~you~are~sad,~you~\textbf{cry}. \\
			\hline
			热 & 流汗 & \textbf{热}的时候会流汗。\\
			hot & sweat & Sweat when it is \textbf{hot}. \\
			\hline
			优酪乳 & 甜 & 优酪乳是\textbf{甜}的。\\
			yogurt & sweet & Yogurt is \textbf{sweet}.\\
			\hline	
		\end{tabular}
		\caption{\label{common-sense} Examples for commonsense knowledge. We replace bold words with [MASK] before input.}
	\end{table*}
\end{CJK*}

\paragraph{Encyclopedia Fact}
For factual information, we present knowledge triples in their sourced natural contexts as in Table \ref{fact}.

\begin{CJK*}{UTF8}{gbsn}
	\begin{table*}
		\centering
		\begin{tabular}{|c|c|c|} 
			\hline
			\multicolumn{1}{|c|}{\textbf{Entity}} & \multicolumn{1}{c|}{\textbf{Relation}} & \multicolumn{1}{c|}{\textbf{Attribute}} \\ 
			\hline
			长尾棕蝠 & 目 & 翼手目 \\
			Long-tailed~brown~bats & order & pterodactyles \\ 
			\hline
			\multicolumn{3}{|l|}{长尾棕蝠是哺乳动物，翼手\textbf{目}、蝙蝠科动物。} \\
			\multicolumn{3}{|l|}{Long-tailed~brown~bats~are~mammals,~pterodactyles,~bats.} \\ 
			\hline
			清香砂锅鸡 & 主要原料 & 酒 \\
			Fragrant~Casserole~Chicken & main~ingredients & wine \\ 
			\hline
			\multicolumn{3}{|l|}{清香砂锅鸡是一道美食，主要原料有鸡、香菇、\textbf{酒}。} \\
			\multicolumn{3}{|l|}{The~main~ingredients~of~Fragrant~Casserole~Chicken~are~chicken,~mushrooms,~and~\textbf{wine}.} \\ 
			\hline
			塔吉克国旗 & 颜色 & 红 \\
			the~national~flag~of~Tajikistan & color & red \\ 
			\hline
			\multicolumn{3}{|l|}{塔吉克国旗，主要颜色是\textbf{红}、白、绿三色。} \\
			\multicolumn{3}{|l|}{The~national~flag~of~Tajikistan~is~mainly~\textbf{red},~white~and~green.} \\
			\hline
		\end{tabular}
		\caption{\label{fact} Example cloze questions of encyclopedia facts. Bold words are replaced by [MASK] for input.}
	\end{table*}
\end{CJK*}

\end{document}